\documentclass[pmlr, hyperfootnotes=false]{jmlr}

\RequirePackage{graphicx}
\usepackage{enumitem}
\usepackage{subcaption}
\usepackage{fancyhdr}
\usepackage{helvet}
\usepackage{rotating}
\usepackage{amsmath}
\usepackage{amsfonts}
\usepackage{amssymb}
\usepackage{tikz}
\usepackage{float}
\usepackage{stmaryrd}
\usepackage{soul}
\usepackage{booktabs}
\usepackage{float}
\usepackage{multirow}
\usepackage{hyperref}
\usepackage{longtable} 
\usepackage{dsfont}
\usepackage{etoc}
\usepackage{minitoc}
\usepackage{tocloft}

\renewcommand \thepart{}
\renewcommand \partname{}
 \usepackage{booktabs}
\usepackage{longtable}
\makeatletter
\def\set@curr@file#1{\def\@curr@file{#1}} 
\makeatother
\usepackage[load-configurations=version-1]{siunitx} 

\newcommand{\xgaof}[1]{\textcolor{blue}{\textbf{feedback}}}

\theorembodyfont{\upshape}
\theoremheaderfont{\scshape}
\theorempostheader{:}
\theoremsep{\newline}

\jmlrvolume{252}
\jmlryear{2024}
\jmlrworkshop{Machine Learning for Healthcare}

\title[Predicting Long-Term Allograft Survival in Liver Transplant Recipients]{Predicting Long-Term Allograft Survival \\ in Liver Transplant Recipients}

\author{\Name{Xiang Gao}
       \Email{xgao@cs.toronto.edu}\\ 
       \addr University of Toronto\\
       Vector Institute
       \AND
       \Name{Michael Cooper}
       \Email{coopermj@cs.toronto.edu}\\ 
       \addr University of Toronto\\
       University Health Network\\
       Vector Institute
       \AND
       \Name{Maryam Naghibzadeh}
       \Email{maryam.naghibzadeh@queensu.ca}\\ 
       \addr Kingston Health Sciences Centre
       \AND
       \Name{Amirhossein Azhie}
       \Email{azhiea@myumanitoba.ca}\\ 
       \addr Max Rady College of Medicine, University of Manitoba
       \AND
       \Name{Mamatha Bhat}
       \thanks{Equal Senior Authorship}
       \Email{mamatha.bhat@uhn.ca}\\ 
       \addr University of Toronto\\
       University Health Network
       \AND
       \Name{Rahul G. Krishnan}
       \footnotemark[1]
       \Email{rahulgk@cs.toronto.edu}\\ 
       \addr University of Toronto\\
       Vector Institute}

\begin{document}

\maketitle

\begin{abstract}
Liver allograft failure occurs in approximately 20\% of liver transplant recipients within five years post-transplant, leading to mortality or the need for retransplantation. Providing an accurate and interpretable model for individualized risk estimation of graft failure is essential for improving post-transplant care. To this end, we introduce the \ul{M}odel for \ul{A}llograft \ul{S}urvival (MAS), a simple linear risk score that outperforms other advanced survival models. Using longitudinal patient follow-up data from the United States (U.S.), we develop our models on 82,959 liver transplant recipients and conduct multi-site evaluations on 11 regions. Additionally, by testing on a separate non-U.S. cohort, we explore the out-of-distribution generalization performance of various models without additional fine-tuning, a crucial property for clinical deployment. We find that the most complex models are also the ones most vulnerable to distribution shifts despite achieving the best in-distribution performance. Our findings not only provide a strong risk score for predicting long-term graft failure but also suggest that the routine machine learning pipeline with only in-distribution held-out validation could create harmful consequences for patients at deployment.

\end{abstract}

\section{Introduction}

Liver transplantation is a vital intervention for patients with end-stage liver diseases, offering a lifeline where often no alternative exists \citep{kuntz2009hepatology}. However, graft failure remains a pivotal challenge for post-transplant care as it is still one of the leading causes of long-term mortality for transplant recipients \citep{watt2010evolution}. Although short-term survival in liver transplant recipients has continued to improve, long-term outcomes for one-year survivors remain unchanged over the last three decades \citep{rana2019nogain}. Therefore, the challenge of accurately identifying long-term risk of graft failure in liver transplant recipients represents a problem with the potential for significant clinical impact.

Unlike in pre-transplant care, in which an established risk score (the MELD; \cite{kamath2001model, kim2021meld}) is used to assess risk and prioritize patients for transplant, post-transplant care lacks a standardized algorithm to assess risk of graft failure. Instead, physicians predominantly rely on their experience and domain expertise to make adjudications on a case-by-case basis. Despite the merits of case-by-case adjudication, there are often benefits associated with standardizing risk assessment in clinical care: the introduction of MELD, for example, is widely credited with substantially reducing liver transplant waitlist mortality in the United States \citep{sacleux2019critical}. This underscores the need for a data-driven approach to predict long-term post-transplant outcomes in liver transplant patients.

There has been tremendous progress in applying machine learning to clinical problems: in the space of solid-organ transplantation, machine learning-based methods have been applied to a variety of settings including donor-recipient matching, waitlist mortality prediction, and post-transplant complication diagnosis \citep{gotlieb2022promise}. Specifically, the use of neural networks in deep learning effectively aids the process of discovering hidden patterns and constructing useful representations of high-dimensional health data when sufficient data is available \citep{lecun_deep_2015, rajkomar2018scalable}. Despite many success stories of black-box models and research efforts on building interpretable models \citep{chen2019looks, chen2020concept}, the drawbacks of most neural networks such as being inherently not interpretable can induce bias and become harmful in high-risk domains like healthcare \citep{lundberg2017unified, lee2021development, si2023interpretabnet}. For example, deep learning models are shown to be capable of accurately predicting self-reported racial identity from corrupted medical images \citep{gichoya2022readingface}. Also, the selection of features was often done less carefully with black-box models because they can often automatically extract predictive representations from all the information available without domain expert oversight, which may allow the model to learn spurious correlation that could lead to adverse consequences for patients \citep{geirhos2020shortcut}. We show that using well-selected clinical biomarkers not only improves interpretability but also leads to better or approximately similar performance when training deep survival models. In addition, the evaluations of machine learning models in healthcare often lack comprehensive comparisons across different model complexity and geographical diversity to understand the generalization capability and clinical utility of these models.

Simple risk scores, unlike many sophisticated deep learning systems, are widely used in different clinical settings to aid the process of decision-making. For patients with end-stage liver disease, transplant centers in North America have relied on the Model for End-Stage Liver Disease (MELD) \citep{kamath2001model,kamath2007model}, MELDNa \citep{biggins2005serum,biggins2006evidence}, and MELD 3.0 \citep{kim2021meld} as a measure of risk to prioritize patients on the waitlist for potential transplant opportunities. Previous research on risk scores for liver graft failure primarily focused on short-term outcomes, particularly early allograft dysfunction (EAD), with models developed to predict early post-transplant survival \citep{pareja2015score, agopian2018evaluation}. Nevertheless, many current clinical risk scores are either human-derived using domain expertise or developed with a limited amount of single-center data and lack a comparison to other deep learning approaches. For instance, the original MELD score was developed on only a set of 231 patients with Cox regression \citep{malinchoc2000model}, and recent evidence suggests that deep learning can potentially improve the current MELD-based system for pre-transplant risk assessment \citep{nagai2022use, cooper2022deep}.

In this work, we employ a large cohort from a publicly available registry of transplant patient data in the United States (U.S.) for model development. We demonstrate a simple and interpretable risk score, Model for Allograft Survival (MAS), based on a set of six post-transplant clinical biomarkers that were identified to have strong predictive power for estimating risks of long-term graft failure. Additionally, several black-box models based on the same covariates are developed to further verify the generalization performance from the small model complexity of MAS. We report the model performance on all regions in the Organ Procurement and Transplantation Network (OPTN) and externally validate the generalization of MAS models by performing direct model evaluations on a private local dataset from Canada without any further model fine-tuning or retraining. Our contributions are as follows:
\begin{enumerate}[leftmargin=*]
    \item We introduce a novel linear risk score for liver transplant recipients that demonstrates consistent and comparable predictive accuracy for long-term graft failure over advanced deep learning models, developed using extensive longitudinal follow-up data from 82,959 patients across multiple regions in the United States.
    \item We thoroughly evaluate the models on multiple OPTN regions and a separate cohort from Canada without model re-tuning to assess in- and out-of-distribution generalization performance, showcasing MAS’s robust performance across diverse patient populations and highlighting its potential for broad clinical deployment in the wild.
    \item We provide critical empirical evidence into the vulnerability of complex models to distribution shifts in clinical settings, underscoring the importance of broader validation practices beyond in-distribution held-out validation and pushing back on the idea that black-box models are generally more performant in healthcare.
\end{enumerate}

\subsection*{Generalizable Insights about Machine Learning in the Context of Healthcare}

Our findings demonstrate the importance of external evaluations of learning algorithms applied in the context of healthcare, and we advocate for a paradigm shift in the machine learning pipeline for healthcare applications to ensure patient safety and model reliability upon deployment. By comparing MAS with advanced models and traditional risk scores across multiple geographical locations, we highlight the importance of robust validation frameworks in assessing model generalizability and performance consistency. The results of superior performance from more complex models on an internal held-out test set cannot serve as the sole evidence for model selection, especially in a healthcare setting where a single model is desired for deployment at multiple sites. This work illustrates that we can achieve model interpretability and generalization with comparable predictive accuracy using traditional statistical approaches for important clinical problems, and it also shows that we urgently need new sample-efficient deep learning methods that carry more suitable inductive bias through, for example, prior domain knowledge as structure embedded in neural networks \citep{chen2024structured, weilbach2023graphically}.

\section{Related Work}

\paragraph{Machine Learning in Survival Analysis.} Survival analysis is an important statistical framework that handles time-to-event outcomes with censored subjects, which is particularly suitable for analyzing failure events that are naturally right-censored, i.e., patients who are lost to follow-ups. Instead of directly predicting the patient outcome within a specific time frame, many of the modeling tools from survival analysis, such as Cox regression \citep{cox1972regression}, can provide individualized risk estimations for patients across different time points in the future. Beyond linear models of survival analysis, random survival forests incorporate tree-based methods into handling censored data \citep{ishwaran2008forest}. Recently, various neural-network-based approaches have been proposed to model survival outcomes. For instance, DeepSurv modified the linear Cox regression model to be parameterized by neural networks \citep{katzman_deepsurv_2018}, and DeepHit uses neural networks to model the discrete distribution of survival times and handles competing risks \citep{lee2018deephit}. In addition, \citet{transformer2021hu} integrated the Transformer architecture \citep{vaswani2017attention} into survival modeling, and we extend it to incorporate longitudinal patient trajectories and refer to it as survival Transformer throughout this paper.

\paragraph{Machine Learning in Liver Transplantation.} The application of machine learning to predict outcomes in liver disease and transplantation has garnered significant interest, leading to a thorough review of existing research \citep{spann2020review, tran2022application, wingfield2020using}. Specifically, in post-transplant outcome prediction, several prior works have focused on long-term or short-term patient mortality prediction ranging from 30-day mortality to 5-year mortality after transplant \citep{liu2020predicting, yasodhara2021identifying, ershoff2020training}. Recently, \citet{nitski2021long} illustrated the use of powerful sequence models like Transformers in predicting causes of mortality for liver transplant recipients. However, when it comes to predicting graft failure after liver transplantation, one of the primary causes of post-transplant mortality, most studies have only considered short-term outcome prediction. For instance, \citet{lau2017machine} developed non-linear predictive models which use pre-transplant information to predict 30-day and 90-day graft failure. Such attempts usually target the difficult problem of donor-recipient matching, in the hopes of allocating the graft to recipients who are least likely to have graft failure after the transplant. In this study, our goal is to enhance post-transplant care by focusing on modeling patient follow-ups, rather than relying only on characteristics identified before the transplant. Although there is a common agreement among these studies that complex models, such as neural networks, usually outperform simpler linear models in prediction accuracy, we challenge this notion. We present a counterexample in our work to demonstrate that more complex models do not always yield consistently better predictions.

\paragraph{Graft Failure Risk Score.} Research on conventional clinical risk scores for graft failure has primarily focused on short-term graft survival, in particular, early allograft dysfunction (EAD). The Model for Early Allograft Function Scoring (MEAF) has been proposed to predict graft survival at 3-month to 12-month follow-ups \citep{pareja2015score}. In a later work, \citet{agopian2018evaluation} developed the Liver Graft Assessment Following Transplantation (L-GrAFT) risk score for estimating 3-month graft failure that outperforms the MEAF score \citep{agopian2021multicenter}. More recently, \citet{avolio2020development} created the Early Allograft Failure Simplified Estimation (EASE) score for the same task, and additional covariates beyond laboratory variables are included such as blood transfusion, MELD, and transplant center volume. The Albumin-Bilirubin (ALBI) grade is also a promising candidate for predicting graft failure which relies on a simple combination of only two variables \citep{johnson2015assessment}. Since there are no clinical risk scores that are directly comparable with MAS in predicting long-term outcomes, we include the ALBI grade, MELD score, and a locally trained MEAF score as baselines, and we provide a quantitative comparison of their utility as potential candidates for predicting long-term graft failure in liver transplant recipients.

A recurring challenge in related literature is the limited size and heterogeneity of study cohorts, with numerous studies relying solely on data from single centers without validating their findings across external datasets. Furthermore, there is an evident gap in these studies concerning the comprehensive evaluation of models across a spectrum of complexities. For instance, \citet{lau2017machine} focused on training non-linear models such as random forests and neural networks locally, comparing them against linear models cited from prior research. In contrast, the body of work on risk scores predominantly favored linear models, often overlooking the potential of non-linear methodologies.

\section{Methods}

\subsection{Dataset and Study Participants}
We develop and validate our approaches on two datasets: the Scientific Registry of Transplant Recipients (SRTR) dataset from the U.S. and a private dataset of liver transplant patients from the University Health Network (UHN) in Canada. The SRTR dataset is the largest transplant database in the U.S., where 9527 liver transplantations were performed and recorded in 2022 \citep{kwong2024optn}. In addition, primarily serving the province of Ontario, UHN has the largest liver transplant program in Canada with almost 200 liver transplantations done each year.

The SRTR dataset comprises both static and time-varying patient information including patient demographics, pre-transplant clinical and laboratory variables, and longitudinal post-transplant updates on the covariates. We only include post-transplant laboratory updates for the UHN dataset since there was no model development on that data; however, the full UHN dataset has little overlap with the SRTR in terms of patient characteristics, which poses a challenge on transferring the model when no covariate selection is done.  For each transplant in the SRTR dataset, the follow-ups are recorded at 6 months, 1 year and then annually after transplantation until the graft recipient dies or is lost to follow-up. The UHN dataset contains initial frequent screenings and then follow-ups every 1-3 months depending on the circumstances of the recipients. Moreover, the SRTR database contains data from many transplant centers that can be further clustered into 11 OPTN regions across the U.S.. The UHN dataset represents a different data distribution as there are many factors such as socioeconomic differences in healthcare between the two countries that will affect the post-transplant outcomes. Hence, a direct test of MAS models on the UHN dataset can serve as strong results on generalization performance of the models.

We identified the dates and outcomes of graft failure for each patient in the SRTR dataset through the recorded failure dates in the database, and the outcome is defined by a list of causes that are also provided in the retrospective data, such as acute rejection or recurrent disease that originated the failure of the transplanted graft. We apply the same definition of graft failure to the UHN database to find corresponding dates of graft failure. In addition, for competing risk analysis with DeepHit, we found patients that had mortality events due to causes other than liver failure since graft failure can no longer happen after such mortality event, and these competing events are considered censored events in other models that do not incorporate competing risks. 

\begin{table}[!ht]
    \centering
    \caption{Details of the SRTR and UHN datasets for model training and validation.}
    \vspace{2mm}
    \begin{tabular}{c  p{1.5cm}  p{1.7cm}  p{2cm}  p{1.43cm}  p{1.5cm} p{2cm} } 
         \hline
          & Total Patients & Observed (\%) & Censored (\%) & Full Features & MAS Features & Max Time \\ [0.5ex] 
         \hline
         SRTR & 82959 & 5259 (6.34\%) & 77700 (93.66\%) & 71 & 6 & 6806 days\\ 
         \hline
         UHN & 3356 & 179 (5.33\%) & 3177 (89.94\%) & N/A & 6 & 12001 days\\ 
         \hline
    \end{tabular}
    \label{table:df_detail}
\end{table}

In both datasets, we kept only the adult patients by excluding patients younger than 18 years at the time of transplantation, and we also excluded patients who have undergone more than one liver transplantation or multiple-organ transplantation. Similarly to \cite{nitski2021long}, patient records between February 2002 and December 2021 from the SRTR database are included in our study due to the incomplete information before 2002, and transplantations conducted from December 1986 to June 2022 are eligible data entries in the UHN dataset. Since this work focuses on the long-term outcome prediction using follow-up data, we additionally removed recipients without any longitudinal follow-up updates. Also, we summarize the cohort selection process with a table of clinical exclusions and corresponding number of samples excluded in Appendix \ref{appendix:cohort_select}. As indicated in Table \ref{table:df_detail}, we included 82959 patients from the SRTR cohort and 3356 patients from the UHN cohort. For details on the specific composition of the feature set used for model development, please refer to Appendix \ref{appendix:feature_choice}. 

\subsection{Models}

We consider five models with distinct modeling assumptions and varied input requirements to benchmark their ability in estimating individualized risk over time: time-varying Cox regression, random survival forest \citep{ishwaran2008forest}, DeepSurv \citep{katzman_deepsurv_2018}, DeepHit \citep{lee2018deephit}, and survival Transformer \citep{transformer2021hu}. We also include a variant of the survival transformer by replacing the model architecture to be recurrent neural networks while keeping the same loss formulations to further assess the gains from modeling the temporal trends. These models will output a survival quantity including the hazard, survival probability, or the cumulative incidence that can be used as a measure of risk for a outcome with censorship.

All models except the survival Transformer and the time-varying Cox model make predictions only based on the most recent patient covariates, whereas the other two models directly work with the longitudinal history of each patient. However, for the time-varying Cox model, longitudinal data were only utilized during training when constructing the partial likelihood, and the model still makes predictions based on the most recent observations. This modeling difference can lead to informative results for determining if temporal trends can provide additional predictive signals. Moreover, we incorporated competing risk analysis with DeepHit since the mortality of recipients due to reasons other than liver failure can prevent the occurrence of graft failure events.

For comparing with other clinical risk scores for early graft dysfunction, we cannot directly implement their original score definitions because we do not have access to the post-operative values that are only days after the transplantation. Instead, we use and compare their choice of biomarker selection to MAS and build a time-varying Cox model with the corresponding feature set. Due to the lack of access to other patient covariates, we only include a comparison to MEAF from the family of risk scores on EAD since the inputs of MAS strictly contain the  biomarkers of MEAF. Although we can directly calculate the ALBI grade and MELD score on our datasets, we still include a comparison with locally trained such models.

Our model development was completed using Python. For implementations of each model, we used \texttt{lifelines} for the time-varying Cox model, \texttt{scikit-survival} for random survival forest, \texttt{pycox} for DeepHit, and the authors' original code repository for DeepSurv and survival Transformer. The code to reproduce the results presented in this work is available on GitHub at \url{https://github.com/rgklab/model_allograft_survival}.

\subsection{Experimental Setup}

\paragraph{Model Training and Validation.} We performed all model development and internal validation on the SRTR dataset due to its large sample size and diversity of regional coverage. On the SRTR dataset, liver transplant recipients were randomly split into training ($70\%$), validation ($15\%$), and test ($15\%$) sets. Hyperparameter tuning was done on the validation set; the chosen optimal hyperparameters and the space searched can be found in Appendix \ref{appendix:hp}. We report the model performance on the test set, and the test set remained unused until the end of model development and for final evaluation. The entire UHN dataset is reserved for external validation until the end of model training on the SRTR dataset, and we report the direct test results of MAS-based models. No fine-tuning or model re-training was done on the UHN dataset. We bootstrapped 1000 samples for all calculations of confidence intervals in this study. Missing data entries in both datasets were forward-filled using observed values from prior follow-up as in \citet{nitski2021long}; the training set mean was used for imputation when there is no past value.

The baseline observation time is defined as the time of transplant, and survival is calculated from the baseline time to the date of relevant event or last follow-up. All models except the survival Transformer and the time-varying Cox model utilized the augmented dataset since their training and predictions only involve the most recent input values instead of the longitudinal trend. The augmented dataset treats longitudinal data of one individual patient as multiple independent samples by setting the time of measurement as baseline and calculating the time-to-event as the difference between event time and the baseline of measurement time. In this case, we makes the simplifying assumption that each patient follow-up observation can be seen as an independent sample though this may be violated since the features might be correlated from prior follow-ups. Additionally, by this formulation, we are making the assumption that patients' future risks of graft failure only depend on current observations, and the past history is irrelevant. Nevertheless, this assumption does not apply to the survival Transformer and its recurrent variant as they are sequence models that take the entire longitudinal history to make predictions.


For initial model development and variable selection, we incorporated 60 static features (240 input variables) and 11 dynamic features (23 input variables) from the SRTR dataset as the full feature set. By training a Cox regression model with elastic net regularization on the full feature set, we aim to identify a small set of strong predictive factors. Subsequently, we trained models with this MAS feature set to construct simpler risk models and evaluate them on the UHN dataset.

\paragraph{Evaluation Metrics.} Similar to most survival analysis studies, we use Harrell's concordance index for evaluating all the models \citep{harrell1982evaluating}. The concordance index measures how accurately a model ranks patients based on the estimated risks, directly reflecting the primary utility of risk models. 

However, to illustrate the model performance at different follow-up times and its ability to estimate long-term risks, we adopt the modified time-dependent concordance index (TDCI) $C(t, \Delta t)$ defined in \citet{lee2019dynamic} to evaluate the performance of our approach. It takes in a prediction time $t$ and a prediction horizon $\Delta t$ to compute a Harrell's concordance index by the risk calculated from covariates at $t$ and limit to the patients who had events within the prediction window $t+\Delta t$. The prediction time $t$ indicates the time at which the prediction is made, and the prediction window $\Delta t$ controls the evaluation window which is the time since the prediction is made. For example, if we want to evaluate the model's performance on predicting $5$-year graft failure using patient follow-ups at year two after transplant, we can calculate $C(t=2, \Delta t=5)$. In the original formulation, \citet{lee2019dynamic} focused on competing risk analysis and added a cause-specific component to the proposed TDCI by treating the competing event as censoring. In our evaluations, we only focus on the performance in terms of the graft failure event, even in the case of DeepHit; nevertheless, the calculation of TDCI on graft failure is the same for all models as we treat any other events as a form of censoring. Details on the formulation of TDCI can be found in Appendix \ref{appendix:eval_metric}.

We mostly consider $t = [0.5, 1, 3, 5]$ and $\Delta t = [1, 3, 5, 7]$ (both in the number of years) which gives us a $4 \times 4$ matrix of concordance index. The prediction time starts from $0.5$ due to the first follow-up data in the SRTR dataset being recorded at 6 months after transplant. For ease of presentation, we average this matrix to a single number and refer to it as mean TDCI in the following sections.

\section{Model Evaluations and Score Construction}

\begin{figure}[!ht]
        \centering
        \includegraphics[width=0.8\linewidth]{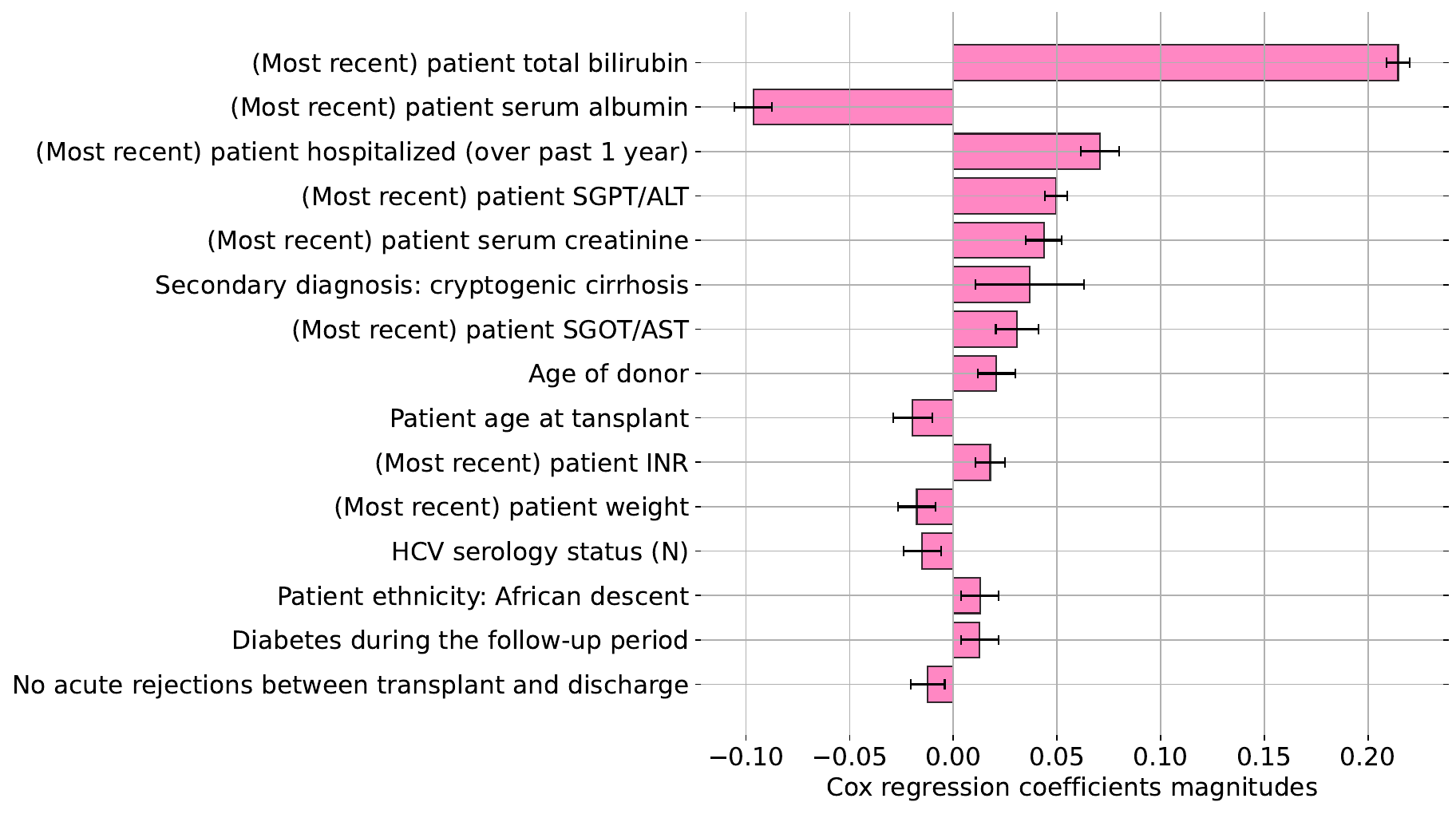}
        \caption{Coefficients from the time-varying Cox model trained with full feature set.}
        \label{fig:feature_importance}
\end{figure}

The initial Cox regression model with all the features as input showed that certain covariates contain the majority of the predictive power. From the ranking of the penalized coefficients as shown by Figure \ref{fig:feature_importance}, we identified the following six laboratory test features with high predictive power: total bilirubin, creatine, albumin, aspartate aminotransferase (AST), alanine aminotransferase (ALT), and international normalized ratio (INR). These biomarkers are among the top influential covariates, and they are commonly available laboratory test results, making the model more broadly accessible and easy to deploy in different sites that may collect different covariates. Here, we refer to this set of input features as MAS feature set and the simple risk score from Cox model trained on the MAS feature set as MAS. Furthermore, in Table \ref{table:SRTR_eval} we demonstrated that this MAS feature set gave approximately comparable results compared to training with the full feature set.

Models were trained with both the full feature set and the MAS feature set, and we refer to the models with MAS feature set as MAS, MASForest, DeepMAS, MASHit, and MASFormer in later sections. Since MAS is only a linear combination of the input biomarkers, it can be directly written as the following equation (units are z-score normalized):
\begin{equation*}
    \textit{MAS}: 20.42\ \text{bilirubin} - 5.69\ \text{albumin} + 2.88\ \text{creatine} + 1.55\ \text{INR} + 5.05\ \text{ALT} + 2.14\ \text{AST}
\end{equation*}

We evaluated all models on the reserved SRTR test dataset with the full feature set as well as the MAS feature set and report the results in Table \ref{table:SRTR_eval}. The random survival forest achieved the highest concordance $0.837$ [$0.829-0.844$] with the full feature set as input covariates, and the DeepHit model obtained higher concordance $0.828$ [$0.821-0.834$] than the other models with the MAS features. Also, both DeepHit and random survival forest showed top performance across the two feature sets with a maximum difference of $0.002$. However, the survival Transformer model along with its recurrent variant capable of learning longitudinal trends performed worse than the other non-linear models with only access to the most recent patient covariates, and this aligns with our intuition that the temporal signal from annual follow-ups is relatively small. Although the Cox model did not achieve the best result, it shows comparable and even better performance to its black-box counterparts within the training distribution. The use of MAS features did not cause much performance degradation, and it enabled faster training with considerably smaller input dimensions (6 v.s. 263); later, we also show that it enables easy model transfer to other transplant centers. In the case of DeepSurv, the model can obtain better in-distribution test performance with the MAS features.

\begin{table}[!ht]
    \centering
    \caption{Model performance (95\% CI) on the SRTR dataset evaluated with mean TDCI, highlighting the best performing model in each feature set.}
    \vspace{3mm}
    \begin{tabular}{c  p{4cm}  p{4cm} }
         \toprule
           & \textsl{Full Feature Set} & \textsl{MAS Feature Set} \\ 
         \midrule
         \textsc{Time-varying Cox} & 0.831 (0.824 - 0.838) & 0.819 (0.812 - 0.826)  \\ 
         \hline
         \textsc{Random Survival Forest} & \textbf{0.837} (0.829 - 0.844) & 0.827 (0.820 - 0.834)\\
         \hline
         \textsc{DeepSurv} & 0.814 (0.807 - 0.821)  & 0.825 (0.819 - 0.833) \\ 
         \hline
         \textsc{DeepHit} & 0.835 (0.828 - 0.842)  & \textbf{0.828} (0.821 - 0.834) \\ 
         \hline
         \textsc{Recurrent Neural Network} & 0.811 (0.804 - 0.819)  & 0.785 (0.777 - 0.793)  \\ 
         \hline
         \textsc{Survival Transformer} & 0.820 (0.813 - 0.827) & 0.819 (0.812 - 0.826) \\ 
         \bottomrule
    \end{tabular}
    \label{table:SRTR_eval}
\end{table}

\section{Geographical Generalization of Survival Models}

As the SRTR dataset contains a collection of transplant centers in the U.S. and can be coherently separated into 11 OPTN regions across the country, we additionally report the model performance under each internal region in the United States which we refer to as in-distribution regions since they were part of the training dataset. Also, with access to an external non-U.S. dataset, we are able to direct test on an out-of-distribution region which highlights the models' generalizability.

We evaluated MAS-based models on all OPTN regions and the UHN dataset, and the results are shown in Table \ref{table:3}. The data on the 11 OPTN regions are from the SRTR reserved test set to ensure no testing samples were leaked into training. Similar to the results on the aggregated data, non-linear models with higher model capacity such as DeepHit perform moderately better on each of the in-distribution region. Among the OPTN regions, all models can achieve above $0.800$ concordance among the regions except for regions 1, 5, and 7, but we can still reach above $0.750$ in these three regions. In the next section, we will showcase through MAS that this slight performance disparity is not introduced through training on the joint dataset but rather exists within the region-specific data.

It is important to note that there was no model fine-tuning or re-training on the UHN dataset as we believe a direct test on out-of-distribution data will serve as stronger evidence for model generalization. Surprisingly, DeepHit, which was the best-performing model on the SRTR dataset $0.828$ [$0.821-0.834$], performed the worst on the UHN dataset $0.749$ [$0.725 - 0.769$]. MAS being the simplest model in terms of model complexity obtained the highest concordance $0.847$ [$0.832 - 0.862$] on the UHN dataset.

\begin{table}[!htb]
    \centering
    \caption{MAS-based model performance (95\% CI) on all OPTN regions and the UHN data evaluated with mean TDCI.}
    \vspace{3mm}
    \resizebox{.95\textwidth}{!}{ 
    \begin{tabular}{c  p{2.5cm}  p{2.5cm}  p{2.5cm}  p{2.5cm}  p{2.5cm}}
         \toprule
           & \textsc{MAS} & \textsc{MASForest} & \textsc{DeepMAS} & \textsc{MASHit} & \textsc{MASformer} \\ 
         \midrule
         \multicolumn{1}{l}{OPTN 1} & 0.811 & 0.824 & 0.822 & 0.819 & 0.818  \\ 
         & (0.682 - 0.840) & (0.697 - 0.852) & (0.691 - 0.852) & (0.691 - 0.850) & (0.697 - 0.846)  \\
         \hline
         \multicolumn{1}{l}{OPTN 2} & 0.842 & 0.847 & 0.853 & 0.856 & 0.842 \\
          & (0.830 - 0.853) & (0.836 - 0.857) & (0.842 - 0.863) & (0.847 - 0.866) & (0.831 - 0.853) \\
         \hline
         \multicolumn{1}{l}{OPTN 3} & 0.833 & 0.834 & 0.845 & 0.840 & 0.830 \\ 
          & (0.821 - 0.844) & (0.822 - 0.845) & (0.834 - 0.854) & (0.828 - 0.849) & (0.820 - 0.841) \\ 
         \hline
         \multicolumn{1}{l}{OPTN 4} & 0.848 & 0.851 & 0.850 & 0.854 & 0.852 \\ 
          & (0.835 - 0.862) & (0.835 - 0.866) & (0.836 - 0.865) & (0.839 - 0.868) & (0.838 - 0.865) \\
         \hline
         \multicolumn{1}{l}{OPTN 5} & 0.774 & 0.789 & 0.781 & 0.785 & 0.799\\ 
          & (0.756 - 0.792) & (0.771 - 0.807) & (0.762 - 0.798) & (0.767 - 0.802) & (0.782 - 0.812) \\
         \hline
         \multicolumn{1}{l}{OPTN 6} & 0.838 & 0.865 & 0.861 & 0.878 & 0.829 \\ 
          & (0.702 - 0.863) & (0.732 - 0.884) & (0.728 - 0.885) & (0.744 - 0.899) & (0.695 - 0.854) \\
         \hline
         \multicolumn{1}{l}{OPTN 7} & 0.811 & 0.819 & 0.820 & 0.827 & 0.815 \\
          & (0.792 - 0.827) & (0.800 - 0.835) & (0.798 - 0.836) & (0.805 - 0.843) & (0.797 - 0.830) \\
         \hline
         \multicolumn{1}{l}{OPTN 8} & 0.860 & 0.862 & 0.864 & 0.870 & 0.854\\ 
          & (0.842 - 0.875) & (0.843 - 0.877) & (0.847 - 0.879) & (0.854 - 0.884) & (0.833 - 0.868) \\
         \hline
         \multicolumn{1}{l}{OPTN 9} & 0.839 & 0.856 & 0.853 & 0.854 & 0.845\\ 
          & (0.735 - 0.857) & (0.745 - 0.870) & (0.749 - 0.868) & (0.749 - 0.869) & (0.743 - 0.861) \\
         \hline
         OPTN 10 & 0.823 & 0.822 & 0.827 & 0.833 & 0.828\\ 
          & (0.802 - 0.840) & (0.803 - 0.840) & (0.807 - 0.845) & (0.814 - 0.850) & (0.808 - 0.844) \\
         \hline
         OPTN 11 & 0.849 & 0.839 & 0.847 & 0.844 & 0.846\\ 
          & (0.831 - 0.864) & (0.819 - 0.855) & (0.829 - 0.864) & (0.826 - 0.862) & (0.828 - 0.862) \\
         \toprule
         \multicolumn{1}{l}{\textbf{UHN}} & 0.847 & 0.840 & 0.829 & 0.749 & 0.776\\ 
          & (0.832 - 0.862) & (0.824 - 0.856) & (0.812 - 0.845) & (0.725 - 0.769) & (0.760 - 0.790) \\ 
         \bottomrule
    \end{tabular}
    }
    \label{table:3}
\end{table}

\section{Benchmarking MAS Against Established Risk Scores}

Furthermore, we systematically evaluate MAS and several related risk scores across different geographical locations and over time. For locally trained risk-scoring models, we denoted them with their names and the prefix "Cox". These models were trained using Cox regression with the same set of input biomarkers as the existing score definitions, allowing us to evaluate the choice of biomarkers. We first evaluated the risk scores in the in-distribution OPTN regions to compare their relative performance and assess whether there were performance disparities resulting from joint training on all regions.

\begin{figure}[!htb]
        \centering
        \includegraphics[width=\linewidth]{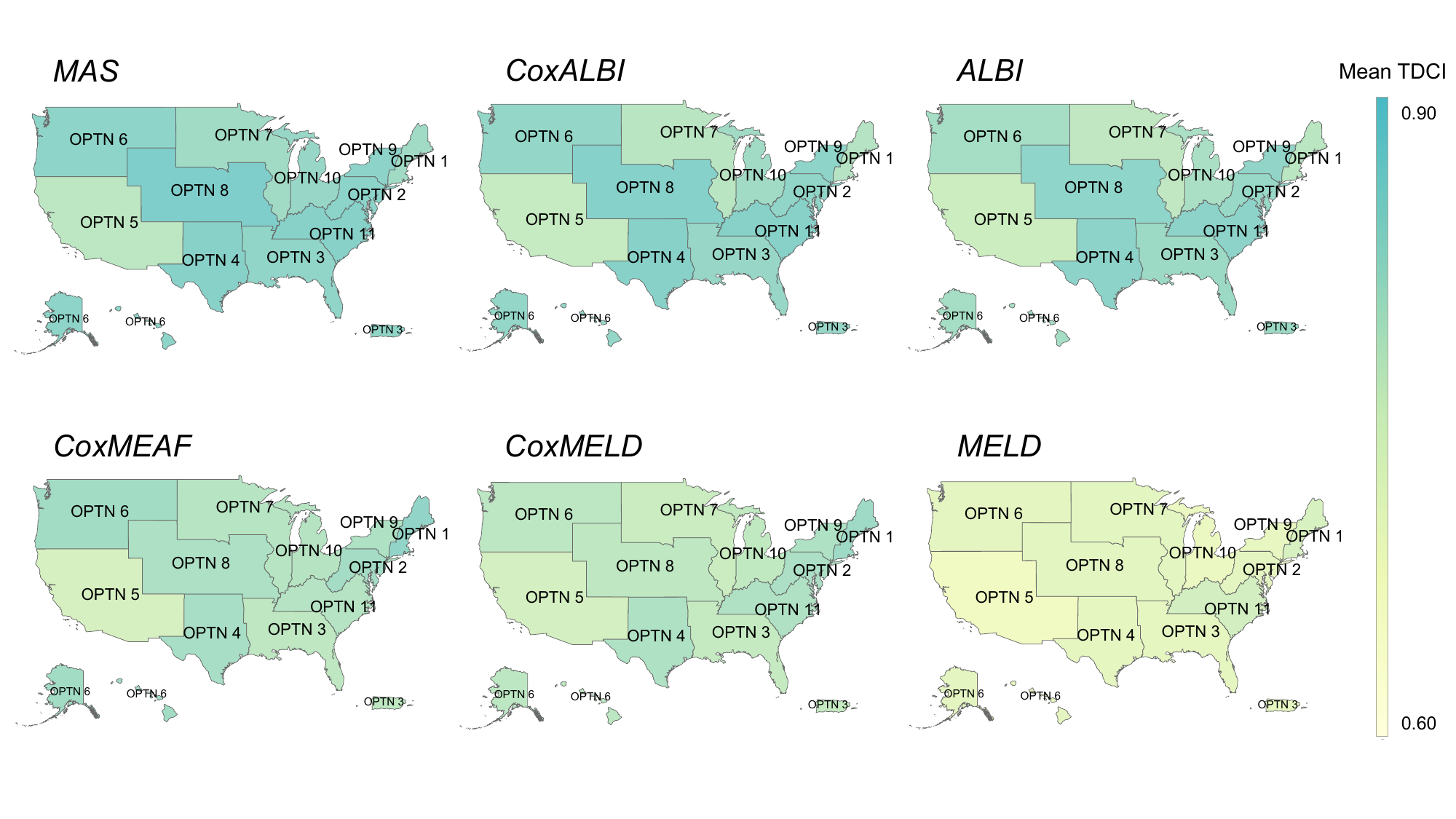}
        \caption{Concordance of risk-scoring models across in-distribution OPTN regions evaluated with mean TDCI. The color intensity indicates the corresponding concordance of the model.}
        \label{fig:map_risk}
\end{figure}

As shown in Figure \ref{fig:map_risk}, the choropleth maps over the United States display the model concordance using color coding to demonstrate in-distribution performance of the models as well as disparities across regions. Cox MEAF, Cox MELD, and particularly MELD exhibit consistently poor concordance across the OPTN regions, whereas MAS obtained the best concordance overall. With the exception of OPTN region 1, MAS is the best-performing model in all other regions, while the MELD score obtains the worst concordance overall. For the exact numbers and full result tables, please refer to Appendix \ref{appendix:eval_risk_scores}.

\begin{figure}[!htb]
        \centering
        \subfigure[]{\includegraphics[width=0.35\linewidth]{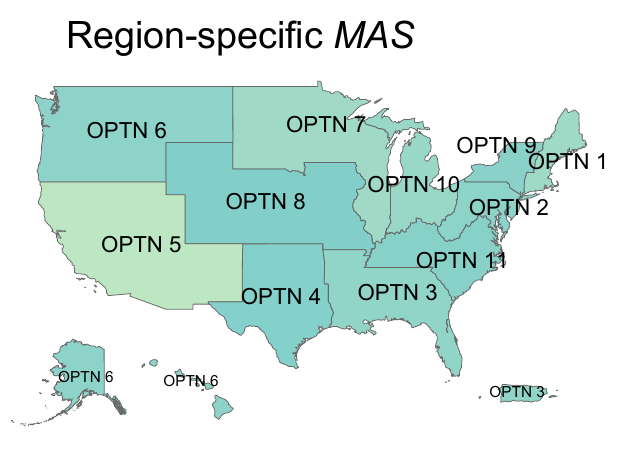}}
        \subfigure[]{\includegraphics[width=0.55\linewidth]{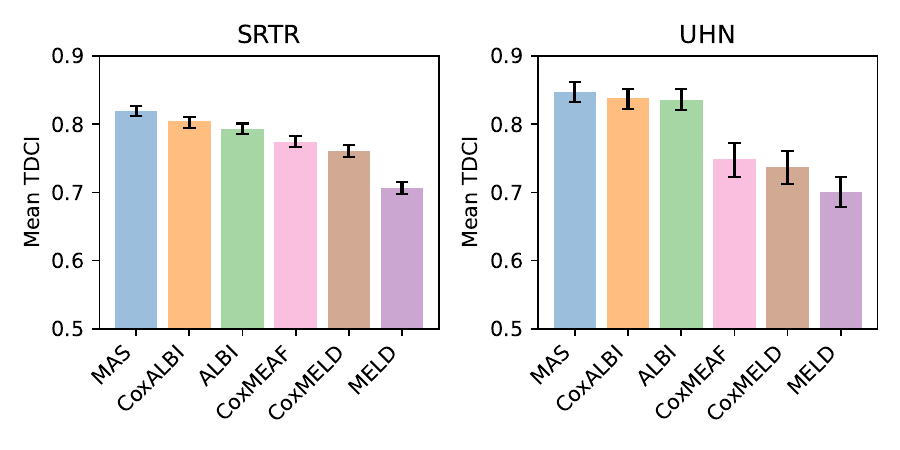}}
        \caption{(a) Concordance of MAS trained on data from each individual region evaluated with mean TDCI. The choropleth map has the same color scale as in the previous figure for comparison. (b) Histograms of risk score performance on both SRTR and the UHN data with 95\% confidence intervals.}
        \label{fig:single_map_hist_risk}
\end{figure}

Our analysis revealed that OPTN region 5 has the worst-case performance for all the risk scores, indicating performance disparities. To investigate these disparities, we trained and evaluated region-specific MAS to measure the optimal performance that the models can achieve on each region using data only from that location. We used the same training process as MAS with data from each region instead of the large joint dataset. This can prevent the models from learning spurious correlations that exist only in a subset of regions and ensure that no regions are disregarded during the training of MAS. As shown in the choropleth map in Figure \ref{fig:single_map_hist_risk} (a), we observed nearly identical performance from the region-specific MAS compared to MAS trained on the entire SRTR, eliminating the possibility of spurious correlations and verifying that all regions were considered during training.

Moreover, we evaluate the generalization performance of the risk scores on the out-of-distribution non-U.S. dataset. As demonstrated in Figure \ref{fig:single_map_hist_risk} (b), MAS outperformed all other potential risk scores in assessing long-term graft failure. In both datasets, MAS achieved the highest mean TDCI with $0.819$ [$0.812 - 0.826$] on the SRTR and $0.847$ [$0.832 - 0.862$] on the UHN dataset. Other scores except for Cox ALBI and ALBI struggled with long-term prediction, particularly on the UHN dataset. Cox ALBI and ALBI came in second and third in predicting graft failures on the two datasets as both albumin and bilirubin are also important predictors in MAS. Furthermore, the performance of locally trained MEAF and MELD also suggested the superiority of the selection of MAS features. It is worth noting that the small model complexity of these risk scores generally showed less performance degradation when generalizing to an out-of-distribution center compared to the neural networks we presented previously.

\begin{figure}[!htb]
        \centering
        \includegraphics[width=\linewidth]{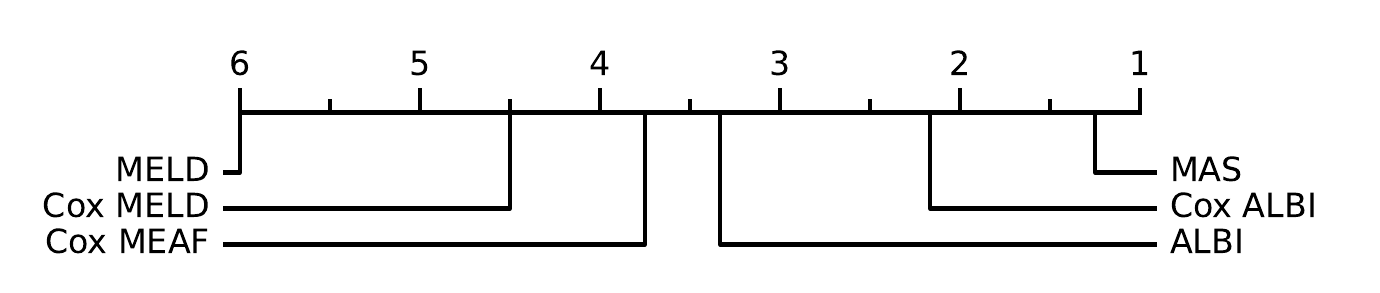}
        \caption{Critical difference diagram showing pairwise statistical difference comparison of the six risk scores on all the OPTN regions as well as the UHN data.}
    \label{fig:4}
\end{figure}

One important tool for comparing multiple models over multiple datasets is the critical difference diagram \citep{demsar2006statistical, benavoli2016should}. Such a diagram concisely represents a two-stage hypothesis testing process. First, there must be a statistical difference among the results determined by the Friedman test. This is followed by pairwise Wilcoxon signed-rank tests, which tell us whether each pair of models exhibits a significant difference in terms of performance. Finally, we adjust for multiple testing with Holm's correction. In our case, we compared six risk scores on twelve distinct datasets comprising the OPTN regions and the non-U.S. source. The corresponding critical difference diagram is shown in Figure \ref{fig:4}. The position of the risk scores represents their mean ranks across all datasets, where low ranks indicate that a model outperforms its competitors more often than those with higher ranks. Two or more models are connected with each other by a thick horizontal line if we cannot tell their performance apart in terms of statistical significance. MAS clearly outperforms all other risk scores with the lowest rank, which is close to one, and the results are significantly different since there is no horizontal line connecting MAS with other scores.

\begin{figure}[!hbt]
        \centering
        \includegraphics[width=\linewidth]{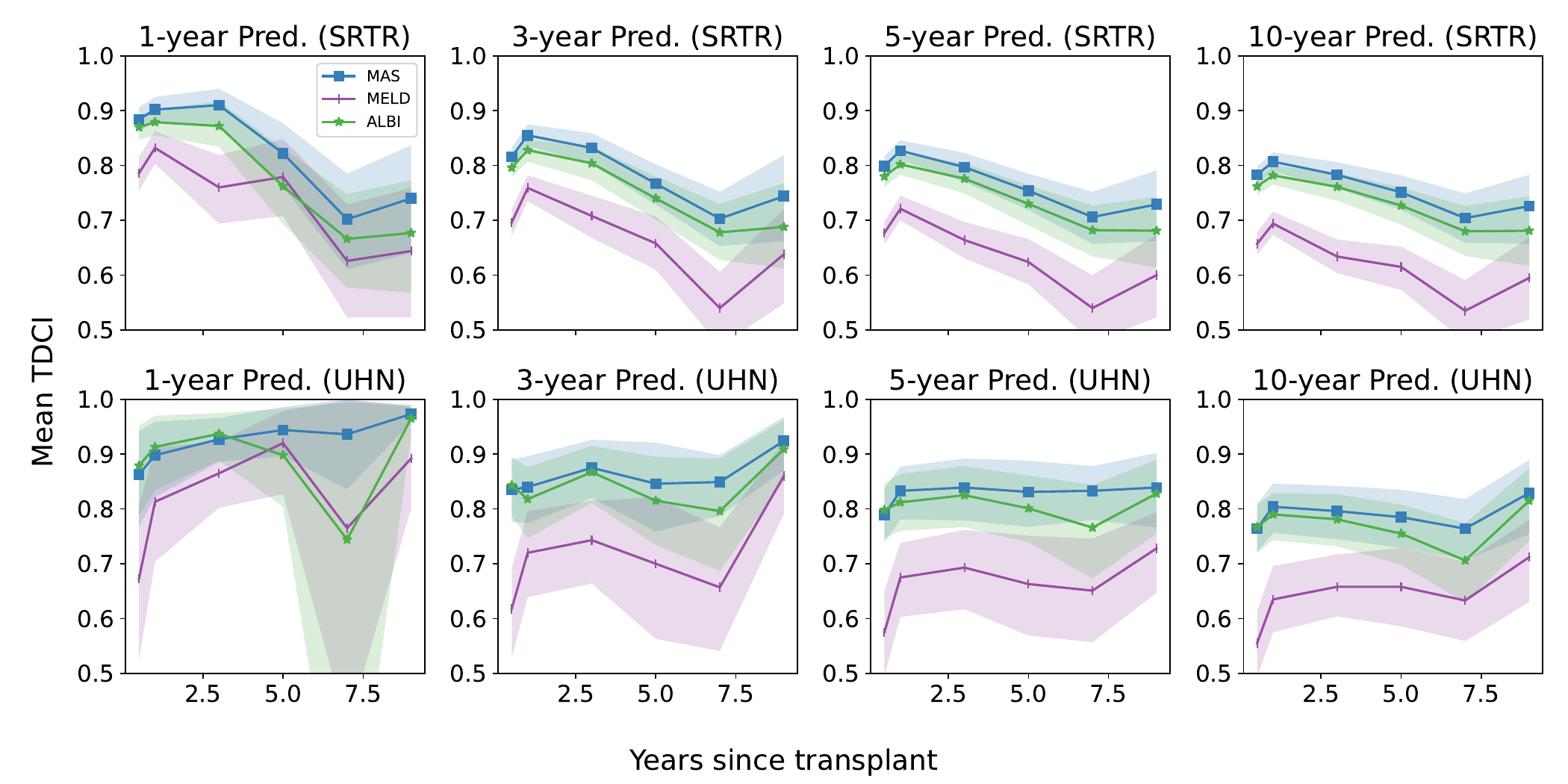}
        \caption{Model concordance on predicting 3-year, 5-year, and 10-year graft failure assessed at different prediction points. The upper row are results on the SRTR dataset, and the bottom row corresponds to the UHN dataset. We bootstrap 1000 samples to obtain the 95\% confidence intervals.}
    \label{fig:risk_line}
\end{figure}

To provide a more comprehensive assessment of the models' predictive capabilities beyond the average performance, we evaluate their performance across different prediction times and windows in Figure \ref{fig:risk_line}. For ease of presentation and clarity, we only include MAS, MELD, and ALBI in the plot; please refer to Appendix \ref{appendix:eval_risk_scores} for a full comparison including all risk scores. We make predictions from various time points since the time of transplant to ensure that the risk scores can be used at any follow-up visits without significant performance degradation. In terms of long-term risk assessment, MAS consistently outperforms the other risk scores in predicting graft failure across both time and geographical locations. Cox ALBI and ALBI perform better than the other scores but exhibit unstable performance, particularly in the UHN evaluations. For example, in the UHN dataset, MAS maintains its performance across multiple time points after transplant for one-year prediction, whereas the other scores show more significant changes in performance.

\section{Discussion} 

We develop an interpretable risk-score to enable monitoring of liver transplant recipients and demonstrate its compelling results in predicting long-term graft failure. We evaluate our risk score against a rich set of model classes as well as related scores on both in- and out-of-distribution datasets. While the MAS model does not strictly outperform all black-box competitors on the SRTR dataset, it provides an excellent tradeoff between simplicity and performance for predicting long-term graft failure.

By selecting a small set of common physiological variables instead of using all the features, we were able to directly test the generalization ability of MAS on a dataset from a distinct transplant center, eliminating the need for fine-tuning or re-training. Our results showed that the simplicity of MAS also enabled it to match and sometimes generalize better to out-of-distribution data than other MAS-based models. This renders it more robust and reliable than its counterparts in generalizing to unseen environments such as a new hospital.

The generalization performance and interpretability of the MAS model make it widely applicable for monitoring graft functioning during follow-ups. It provides risk measures based on retrospective data to aid physicians in the decision-making process for preventative treatment. Larger and more complex models, such as Transformers and random forests, may outperform MAS in some cases, they typically have intensive computation requirements and cannot be easily incorporated into an online calculator. These models may not be accessible to sites with limited computational resources and small patient populations. Complex models may also be prone to be vulnerable against distribution shifts, as shown in our empirical evidence. We would also like to highlight that there are different levels of generalization that can be suitable for specific use cases of learning algorithms in healthcare, and how it relates to the evaluations of models is extensively discussed in the literature \citep{youssef2023external, futoma2020myth}. Direct external validation may not be the gold standard for a model that is only considered for deployment within a single site. Nevertheless, external evaluations can still be beneficial in verifying the robustness of complex models since they can easily rely on spurious information to make predictions.

\paragraph{Limitations} MAS can only rank relative risks with each patient; it does not allow for causal interpretations of its parameters and outputs without further assumptions. It is not designed to provide actionable inference for specific treatments or interventions, and clinicians should exercise caution when using MAS to recommend interventions for patients. Also, MAS does not account for certain forms of bias that may be present in the survival data, such as dependent censoring \citep{gharari2023copula}. 

\acks{We acknowledge the financial support of the Data Sciences Institute at the University of Toronto. This research is also supported by a Canadian Institutes of Health Research (CIHR) Bridge Grant. MC is supported by a CIHR Health Systems Impact Fellowship and by the Vector Institute. Resources used in preparing this research were provided, in part, by the Province of Ontario, the Government of Canada through CIFAR, and companies sponsoring the Vector Institute.}

\newpage
\bibliography{reference}

\newpage
\appendix
\addcontentsline{toc}{section}{Appendix} 
\part{Appendix} 
\parttoc 
\section{Data and Code Availability}

\subsection{Data Availability}
The SRTR dataset on transplant patients from the U.S. is publicly available, and research data access for SRTR Standard Analysis Files can be requested at \href{https://www.srtr.org/requesting-srtr-data/data-requests}{srtr-data-requests} with cost under signed data use agreement. The UHN patient data used in this work is, unfortunately, not open to public research use for privacy and safety reasons.

This study and the use of patient data were reviewed and approved by the OPTN as well as the Research Ethics Board at the University Health Network.

\subsection{Code Availability}
The code for data processing and reproducing the experiments presented is available on GitHub at \url{https://github.com/rgklab/model_allograft_survival}.

\section{Cohort and Dataset Processing}

\subsection{Cohort Selection}
\label{appendix:cohort_select}

In Table \ref{tab:clinical_exclusions_srtr} and \ref{tab:clinical_exclusions_non}, we illustrate the detailed process of applying the clinical exclusions to arrive at the final cohort used in our study for both datasets.

\begin{table}[!ht]
    \centering
    \caption{Complete set of the inclusion/exclusion criteria for the SRTR data, reflecting any clinical exclusions during the selection of the patient cohort for model development.}
    \begin{tabular}{|p{6cm}|c|p{6cm}|c|}
    \hline
     Exclusion Criterion & $N_{\text{excluded}}$ & Remaining Cohort & $N_{\text{remaining}}$\\
     \hline\hline
     Initial Cohort & - & All Transplanted Patients Listed from December 11, 1985 to February 27, 2021. & 183721 \\
     \hline
     Remove patients listed outside time of interest (Feb 27, 2002 – Dec 1, 2021). &    57372 & Patients Listed between Feb 27, 2002 – Dec 1, 2021. &   126349 \\
     \hline
     Remove patients with one or more previous transplants. & 8370 & Patients without previous transplants. & 117979\\
    \hline
     Remove patients who received multi-organ transplants. & 9507 & Patients who received single-liver transplants only. & 108472\\
     \hline
     Remove non-adult patients (under 18 years of age). & 8471 & Adult patients only. & 100001\\
     \hline
     Remove patients ineligible for model training (e.g., without any follow-up information). & 17042 & Patients who are eligible for model development with relevant records. & 82959\\
     \hline
     & & \textbf{Total Included Cohort} & 82959\\
     \hline
    \end{tabular}
    \label{tab:clinical_exclusions_srtr}
\end{table}

\begin{table}[!ht]
    \centering
    \caption{Complete set of the inclusion/exclusion criteria for the UHN data, reflecting any clinical exclusions during the selection of the patient cohort for model development.}
    \begin{tabular}{|p{6cm}|c|p{6cm}|c|}
    \hline
     Exclusion Criterion & $N_{\text{excluded}}$ & Remaining Cohort & $N_{\text{remaining}}$\\
     \hline\hline
     Initial Cohort & - & All Transplanted Patients Listed from December 1, 1986 to June 30, 2022. & 4468 \\
     \hline
     Remove patients with one or more previous transplants. & 505 & Patients without previous transplants. & 3963 \\
    \hline
     Remove patients who received multi-organ transplants. & 172 & Patients who received single-liver transplants only. & 3791 \\
     \hline
     Remove non-adult patients (under 18 years of age). & 239 & Adult patients only. & 3552 \\
     \hline
     Remove patients ineligible for model training (e.g., without any follow-up information). & 196 & Patients who are eligible for model development with relevant records. & 3356 \\
     \hline
     & & \textbf{Total Included Cohort} & 3356 \\
     \hline
    \end{tabular}
    \label{tab:clinical_exclusions_non}
\end{table}

\subsection{Feature Choices and Extraction}
\label{appendix:feature_choice}

Here, we present the features associated with model training for the SRTR dataset in Table \ref{tab:feature_choice}. For full reproducibility, each of the following feature name correspond to the name defined in the SRTR Standard Analysis File. For detailed information associated with each feature name here, please refer to the official data dictionary: \href{https://www.srtr.org/requesting-srtr-data/saf-data-dictionary/}{srtr-data-dictionary}. 

\begin{longtable}{llll}
\caption{Detailed breakdown of the full feature set in SRTR used for model development}\label{tab:feature_choice} \\
\toprule
Feature Name          & Feature Type & Temporal Nature & Missing Rate  \\
\midrule
\endfirsthead

\multicolumn{4}{c}%
{{\bfseries \tablename\ \thetable{} -- continued from previous page}} \\
\toprule
Feature Name          & Feature Type & Temporal Nature & Missing Rate  \\
\midrule
\endhead

\midrule
\multicolumn{4}{r}{{Continued on next page}} \\
\midrule
\endfoot

\bottomrule
\endlastfoot

CANHX\_MPXCPT\_HCC\_APPROVE\_IND             & Binary        & Static  &  0.00\%    \\
CAN\_ANGINA & Categorical & Static &   89.75\%   \\
CAN\_ANGINA\_CAD & Categorical & Static &  52.14\%    \\
CAN\_BACTERIA\_PERIT & Binary & Static &   0.00\%   \\
CAN\_CEREB\_VASC & Binary & Static &  40.45\%    \\
CAN\_DIAB\_TY & Categorical & Static &   0.00\%   \\
CAN\_DRUG\_TREAT\_COPD & Binary & Static &  41.16\%    \\
CAN\_DRUG\_TREAT\_HYPERTEN & Binary & Static &   41.02\%   \\
CAN\_EDUCATION & Categorical & Static &  0.00\%    \\
CAN\_ETHNICITY\_SRTR & Binary & Static &  0.00\%   \\
CAN\_GENDER & Binary & Static &  0.00\%    \\
CAN\_HGT\_CM & Continuous & Static &   0.23\%   \\
CAN\_LAST\_ALBUMIN & Continuous & Static &  0.00\%    \\
CAN\_LAST\_ASCITES & Continuous & Static &   0.00\%   \\
CAN\_LAST\_BILI & Continuous & Static &   0.00\%   \\
CAN\_LAST\_DIAL\_PRIOR\_WEEK & Binary & Static &  0.00\%   \\
CAN\_LAST\_ENCEPH & Continuous & Static &   0.00\%   \\
CAN\_LAST\_INR & Continuous & Static &   0.00\%   \\
CAN\_LAST\_SRTR\_LAB\_MELD & Continuous & Static &  0.00\%   \\
CAN\_LAST\_SERUM\_SODIUM & Continuous & Static &  9.61\%   \\
CAN\_LAST\_STAT & Categorical & Static &  0.00\%   \\
CAN\_MALIG & Binary & Static &   0.00\%   \\
CAN\_PERIPH\_VASC & Binary & Static &  40.53\%   \\
CAN\_PORTAL\_VEIN & Binary & Static &  0.00\%   \\
CAN\_PREV\_ABDOM\_SURG & Binary & Static &  0.00\%   \\
CAN\_PULM\_EMBOL & Binary & Static &  41.35\%   \\
CAN\_RACE\_SRTR & Categorical & Static &   0.00\%   \\
CAN\_TIPSS & Binary & Static &  0.00\%   \\
CAN\_WGT\_KG & Continuous & Static &  0.15\%   \\
DON\_AGE & Continuous & Static &  0.00\%   \\
DON\_TY & Binary & Static &   0.00\%   \\
DON\_WARM\_ISCH\_TM\_MINS & Continuous & Static &  82.48\%   \\
DON\_WGT\_KG & Continuous & Static &   0.08\%   \\
REC\_ACUTE\_REJ\_EPISODE & Categorical & Static &   6.56\%   \\
REC\_AGE\_AT\_TX & Continuous & Static &   0.00\%   \\
REC\_BMI & Continuous & Static &   2.15\%   \\
REC\_CMV\_STAT & Categorical & Static &   1.80\%   \\
REC\_COLD\_ISCH\_TM & Continuous & Static &   3.09\%   \\
REC\_DGN & Categorical & Static &   0.01\%   \\
REC\_DGN2 & Categorical & Static &   0.42\%   \\
REC\_DISCHRG\_SGPT & Continuous & Static &  42.82\%   \\
REC\_FUNCTN\_STAT & Continuous & Static &   0.00\%   \\
REC\_EBV\_STAT & Categorical & Static &   0.94\%   \\
REC\_HBV\_ANTIBODY & Categorical & Static &  1.92\%   \\
REC\_HBV\_SURF\_ANTIGEN & Categorical & Static &  1.92\%   \\
REC\_HCV\_STAT & Categorical & Static &   0.70\%   \\
REC\_IMMUNO\_MAINT\_MEDS & Binary & Static &   0.00\%   \\
REC\_LIFE\_SUPPORT & Binary & Static &   0.00\%   \\
REC\_LIFE\_SUPPORT\_OTHER & Binary & Static &   0.00\%   \\
REC\_MED\_COND & Categorical & Static &   0.00\%   \\
REC\_PORTAL\_VEIN & Binary & Static &   0.00\%   \\
REC\_POSTX\_LOS & Continuous & Static &   0.08\%   \\
REC\_PREV\_ABDOM\_SURG & Binary & Static &   0.00\%   \\
REC\_PRIMARY\_PAY & Categorical & Static &   0.00\%   \\
REC\_TX\_PROCEDURE\_TY & Categorical & Static &   0.00\%   \\
REC\_VENTILATOR & Binary & Static &   0.00\%   \\
REC\_WARM\_ISCH\_TM & Continuous & Static &  49.13\%   \\
REC\_WORK\_INCOME & Binary & Static &  6.46\%   \\
TFL\_CREAT & Continuous & Time-varying &   5.66\%   \\
TFL\_INR & Continuous & Time-varying &   87.35\%   \\
TFL\_TOT\_BILI & Continuous & Time-varying &  36.79\%   \\
TFL\_SGPT & Continuous & Time-varying &   69.22\%   \\
TFL\_ALBUMIN & Continuous & Time-varying &  70.39\%   \\
TFL\_SGOT & Continuous & Time-varying &  98.86\%   \\
TFL\_BMI & Continuous & Time-varying &  81.31\%   \\
TFL\_WGT\_KG & Continuous & Time-varying &  76.78\%   \\
TFL\_HOSP & Binary & Time-varying &  35.42\%   \\
TFL\_DIAB\_DURING\_FOL & Binary & Time-varying &   36.18\%   \\
TFL\_PRIMARY\_PAY & Categorical & Time-varying &  36.57\%   \\
\end{longtable}

\section{Model Training and Evaluation}

\subsection{Hyperparameter Search}
\label{appendix:hp}

The following numbered list shows the hyperparameter space that we searched for each model class. We performed grid search over each of the hyperparameter space. Table \ref{table:best_hp} shows the chosen setup for each feature set with respect to the optimal validation TDCI.
 
\begin{enumerate}[leftmargin=*]
    \item \textbf{Cox Regression}
    
    The space of hyperparameter is the following:
    \begin{itemize}
        \item penalizer: \{0, 0.01, 0.1, 0.2, 0.5, 1, 10\}
        \item l1\_ratio: \{0, 0.1, 0.3, 0.5, 0.7, 0.9\}
    \end{itemize}

    \item \textbf{DeepSurv}
    
    The space of hyperparameter is the following:
    \begin{itemize}
        \item num\_layers: \{2, 3, 4\}
        \item num\_nodes: \{16, 64, 128, 512\}
        \item learning\_rate: \{1e-4, 5e-4, 1e-3\}
        \item L2\_reg: \{0, 0.5, 1, 10\}
        \item dropout: \{0, 0.1, 0.3\}
    \end{itemize}

    \item \textbf{Random Survival Forest}
    
    The space of hyperparameter is the following:
    \begin{itemize}
        \item n\_estimators: \{500\}
        \item min\_samples\_split: \{5, 13, 20\}
        \item max\_depth: \{3, 6, 9\}
    \end{itemize}

    \item \textbf{DeepHit}
    
    The space of hyperparameter is the following:
    \begin{itemize}
        \item alpha: \{0.2, 0.5, 0.8, 0.9\}
        \item batch\_size: \{32, 64, 128, 256\}
        \item dropout: \{0, 0.1, 0.3, 0.5\}
        \item layers\_indiv: \{1, 2, 3\}
        \item layers\_shared: \{1, 2, 3\}
        \item learning\_rate: \{1e-2, 1e-3, 5e-3, 5e-4\}
        \item nodes\_indiv: \{16, 64, 128, 256, 512\}
        \item nodes\_shared: \{16, 64, 128, 256, 512\}
        \item sigma: \{0.1, 0.3, 0.5\}
    \end{itemize}

    \item \textbf{Recurrent Neural Network}
    
    The space of hyperparameter is the following:
    \begin{itemize}
        \item num\_layers: \{1, 2, 3, 4\}
        \item hidden\_dim: \{128, 256, 512\}
        \item dropout: \{0, 0.1, 0.3 0.5\}
        \item learning\_rate: \{1e-4, 5e-4, 1e-3\}
        \item batch\_size: \{8, 16, 32, 64\}
        \item loss\_weight: \{0.01, 0.1, 0.5, 1\}
    \end{itemize}

    \item \textbf{Survival Transformer}
    
    The space of hyperparameter is the following:
    \begin{itemize}
        \item num\_attention\_layers: \{1, 2, 3, 4\}
        \item num\_heads: \{1, 2, 4, 8\}
        \item embedding\_dim: \{256, 512\}
        \item dropout: \{0, 0.1, 0.3 0.5\}
        \item learning\_rate: \{1e-4, 5e-4, 1e-3\}
        \item batch\_size: \{8, 16, 32, 64\}
        \item loss\_weight: \{0.01, 0.1, 0.5, 1\}
    \end{itemize}
\end{enumerate}

\begin{table}[!ht]
    \centering
    \caption{Best hyperparameters of all models with two different feature sets.}
    \vspace{6pt}
    \begin{tabular}{p{3cm}  p{6cm}  p{6cm} }
         \hline
          \textbf{Model} & \textsl{Full Feature Set} & \textsl{MAS Feature Set} \\ 
         \hline
         \textsc{Time-varying Cox} & penalizer: 0.1, l1\_ratio: 0 & penalizer: 0.1, l1\_ratio: 0.5  \\ 
         \hline
         \textsc{DeepSurv} & num\_layers: 2, num\_nodes: 512, learning\_rate: 1e-4, L2\_reg: 10, dropout: 0.3 & 
         num\_layers: 3, num\_nodes: 16, learning\_rate: 5e-4, L2\_reg: 1, dropout: 0\\
         \hline
         \textsc{Random Survial Forest} & n\_estimators: 500, min\_samples\_split: 13, max\_depth: 9 &  
         n\_estimators: 500, min\_samples\_split: 20, max\_depth: 9 \\ 
         \hline
         \textsc{DeepHit} & alpha: 0.2, batch\_size: 256, dropout: 0.5, layers\_indiv: 1, layers\_shared: 2, learning\_rate: 5e-3, nodes\_indiv: 16, nodes\_shared: 16 sigma: 0.1 & alpha: 0.5, batch\_size: 256, dropout: 0, layers\_indiv: 1, layers\_shared: 2, learning\_rate: 1e-2, nodes\_indiv: 128, nodes\_shared: 128, sigma: 0.1 \\ 
         \hline
         \textsc{Recurrent Neural Network} & num\_layers: 3, hidden\_dim: 128, dropout: 0.5, learning\_rate: 1e-3, batch\_size: 16, loss\_weigh: 0.01 & 
         num\_layers: 1, hidden\_dim: 256, dropout: 0.5, learning\_rate: 5e-4, batch\_size: 8, loss\_weigh: 0.01\\ 
         \hline
         \textsc{Survival Transformer} & num\_attention\_layers: 3, num\_heads: 2, embedding\_dim: 256, dropout: 0.1, learning\_rate: 1e-4, batch\_size: 64, loss\_weigh: 0.01 & num\_attention\_layers: 3, num\_heads: 1, embedding\_dim: 256, dropout: 0, learning\_rate: 1e-4, batch\_size: 64, loss\_weigh: 0.1 \\ 
         \hline
    \end{tabular}
    \label{table:best_hp}
\end{table}

\subsection{Evaluation Metrics}
\label{appendix:eval_metric}

In this part, we will describe the common Harrell’s concordance index first and show how it can be extended into time-dependent concordance index (TDCI) as in \citet{lee2019dynamic}.

Harrell’s concordance is simply a ratio of the size of two sets. It is based on the assumption that people who live longer should have been assigned lower risk than people who had a shorter life. Hence, all pairs that can be compared under this assumption will require the person who has shorter time is uncensored. We refer to the set of all such pairs as the set of valid pairs $P_v$. Then, all pairs within $P_v$ that have the correct ranking of risks will be in the set of concordant pairs $P_c$. In addition, we use $f(X^{(i)}) \in \mathbb{R}$ to denote the scalar risk score that the model assigns to patient $i$ and $E^{(i)}$ as the event indicator.

Therefore, $P_v$ and $P_c$ can be defined as follows:
\begin{align*}
P_v &= \big\{(i, j) \,\big\vert\, T^{(i)} < T^{(j)},\, E^{(i)} = 1\big\}\\
P_c &= \big\{(i, j) \,\big\vert\, (i, j) \in P_v,\, f(X^{(i)}) > f(X^{(j)}) \big\}\\
\end{align*}
and we can calculate the C-index from the two sets:
\begin{align*}
C_{\text{Harrell}} &= \frac{\left\vert P_c\right\vert}{\left\vert P_v\right\vert}
\end{align*}
Now, to incorporate the concept of the prediction time $t$ and prediction window $\Delta t$, we use $f(X^{(i)}_t, t, \Delta t) \in \mathbb{R}$ to denote the risk score that the model assigns to patient $i$ with feature at time $t$, and this risk model could potentially depend on the time elapsed into the future $\Delta t$. Hence, we can have the following set definitions that depend on $t$ and $\Delta t$:

\begin{align*}
P_v(t, \Delta t) &= \big\{(i, j) \,\big\vert\, T^{(i)} < T^{(j)},\, T^{(i)} < t+\Delta t,\, E^{(i)} = 1\big\}\\
P_c(t, \Delta t) &= \big\{(i, j) \,\big\vert\, (i, j) \in P_v(t, \Delta t),\, f(X^{(i)}_t, t, \Delta t) > f(X^{(j)}_t, t, \Delta t) \big\}\\
\end{align*}

Similarly, the TDCI can be calculated by taking the ratio of the cardinality of two sets above.
\begin{align*}
\text{TDCI}(t, \Delta t) &= \frac{\left\vert P_c(t, \Delta t)\right\vert}{\left\vert P_v(t, \Delta t)\right\vert}
\end{align*}

\section{Additional Results}

\subsection{Evaluations of Risk Scores}
\label{appendix:eval_risk_scores}

Here, we show the complete table of results that were previously summarized in Figure \ref{fig:map_risk} and \ref{fig:single_map_hist_risk}. In addition, we also show the performance plot of all baseline risk scores for long-term prediction.

\begin{table}[!ht]
    \centering
    \caption{Score performances (95\% CI) on the SRTR dataset and the UHN dataset evaluated with mean TDCI. Cox prefix denotes locally trained models with the corresponding biomarker selection of existing risk scores.}
    \vspace{3mm}
    \begin{tabular}{c  p{4cm}  p{4cm} }
         \toprule
           & \textsl{SRTR} & \textsl{UHN} \\ 
         \midrule
         \textsc{MAS} & 0.819 (0.812 - 0.826) & 0.847 (0.832 - 0.862)  \\ 
         \hline
         \textsc{MELD} & 0.706 (0.697 - 0.715) & 0.701 (0.678 - 0.723)\\
         \hline
         \textsc{ALBI} & 0.793 (0.785 - 0.801)  & 0.836 (0.821 - 0.851) \\ 
         \hline
         \textsc{Cox MEAF} & 0.774 (0.766 - 0.783)  & 0.749 (0.722 - 0.773) \\ 
         \hline
         \textsc{Cox MELD} & 0.761 (0.752 - 0.769) & 0.737 (0.712 - 0.761) \\ 
         \hline
         \textsc{Cox ALBI} & 0.804 (0.795 - 0.811)  & 0.838 (0.822 - 0.852) \\ 
         \bottomrule
    \end{tabular}
    \label{tabel:score_two_set}
\end{table}

\begin{table}[!ht]
    \centering
    \caption{Risk score model performances (95\% CI) on different regions evaluated with mean TDCI.}
    \vspace{3mm}
    \resizebox{.95\textwidth}{!}{ 
    \begin{tabular}{c  p{2.5cm}  p{2.5cm}  p{2.5cm}  p{2.5cm}  p{2.5cm} p{2.5cm}}
         \toprule
           & \textsc{MAS} & \textsc{ALBI} & \textsc{MELD} & \textsc{CoxMEAF} & \textsc{CoxMELD} & \textsc{CoxALBI}\\ 
         \midrule
         \multicolumn{1}{l}{OPTN 1} & 0.811 & 0.778 & 0.730 & 0.832 & 0.815 & 0.785  \\
         & (0.682 - 0.840) & (0.657 - 0.810) & (0.600 - 0.761) & (0.708 - 0.860) & (0.684 - 0.845) & (0.661 - 0.819) \\
         \hline
         \multicolumn{1}{l}{OPTN 2} & 0.842 & 0.822 & 0.702 & 0.808 & 0.797 & 0.833\\
          & (0.830 - 0.853) & (0.809 - 0.834) & (0.685 - 0.718) & (0.795 - 0.822) & (0.783 - 0.811) & (0.819 - 0.845)\\
         \hline
         \multicolumn{1}{l}{OPTN 3} & 0.833 & 0.816 & 0.698 & 0.773 & 0.763 & 0.823\\ 
          & (0.821 - 0.844) & (0.803 - 0.827) & (0.681 - 0.715) & (0.759 - 0.786) & (0.749 - 0.777) & (0.811 - 0.833)\\ 
         \hline
         \multicolumn{1}{l}{OPTN 4} & 0.848 & 0.839 & 0.697 & 0.803 & 0.793 & 0.848\\ 
          & (0.835 - 0.862) & (0.824 - 0.852) & (0.673 - 0.721) & (0.784 - 0.821) & (0.774 - 0.811) & (0.834 - 0.862)\\
         \hline
         \multicolumn{1}{l}{OPTN 5} & 0.774 & 0.755 & 0.673 & 0.734 & 0.740 & 0.762\\ 
          & (0.756 - 0.792) & (0.734 - 0.774) & (0.652 - 0.691) & (0.715 - 0.753) & (0.722 - 0.756) & (0.741 - 0.783)\\
         \hline
         \multicolumn{1}{l}{OPTN 6} & 0.838 & 0.804 & 0.702 & 0.809 & 0.774 & 0.828\\ 
          & (0.702 - 0.863) & (0.674 - 0.827) & (0.560 - 0.740) & (0.676 - 0.835) & (0.637 - 0.807) & (0.693 - 0.854)\\
         \hline
         \multicolumn{1}{l}{OPTN 7} & 0.811 & 0.771 & 0.706 & 0.782 & 0.757 & 0.779\\
          & (0.792 - 0.827) & (0.741 - 0.796) & (0.679 - 0.730) & (0.761 - 0.800) & (0.731 - 0.781) & (0.749 - 0.803)\\
         \hline
         \multicolumn{1}{l}{OPTN 8} & 0.860 & 0.834 & 0.706 & 0.795 & 0.773 & 0.851\\ 
          & (0.842 - 0.875) & (0.814 - 0.849) & (0.678 - 0.730) & (0.757 - 0.824) & (0.737 - 0.802) & (0.832 - 0.866)\\
         \hline
         \multicolumn{1}{l}{OPTN 9} & 0.839 & 0.832 & 0.695 & 0.793 & 0.793 & 0.837\\ 
          & (0.735 - 0.857) & (0.804 - 0.850) & (0.607 - 0.722) & (0.695 - 0.818) & (0.691 - 0.814) & (0.816 - 0.854)\\
         \hline
         OPTN 10 & 0.823 & 0.799 & 0.685 & 0.783 & 0.768 & 0.811\\ 
          & (0.802 - 0.840) & (0.775 - 0.823) & (0.660 - 0.706) & (0.758 - 0.804) & (0.742 - 0.789) & (0.784 - 0.832)\\
         \hline
         OPTN 11 & 0.849 & 0.839 & 0.742 & 0.784 & 0.792 & 0.849\\ 
          & (0.831 - 0.864) & (0.823 - 0.855) & (0.722 - 0.761) & (0.762 - 0.805) & (0.773 - 0.810) & (0.832 - 0.864)\\
         \toprule
         \multicolumn{1}{l}{\textbf{UHN}} & 0.847 & 0.836 & 0.701 & 0.749 & 0.737 & 0.838\\ 
          & (0.832 - 0.862) & (0.821 - 0.851) & (0.678 - 0.723) & (0.722 - 0.773) & (0.712 - 0.761) & (0.822 - 0.852)\\ 
         \bottomrule
    \end{tabular}
    }
    \label{table:score_optn}
\end{table}

\begin{figure}[!hbt]
        \centering
        \includegraphics[width=\linewidth]{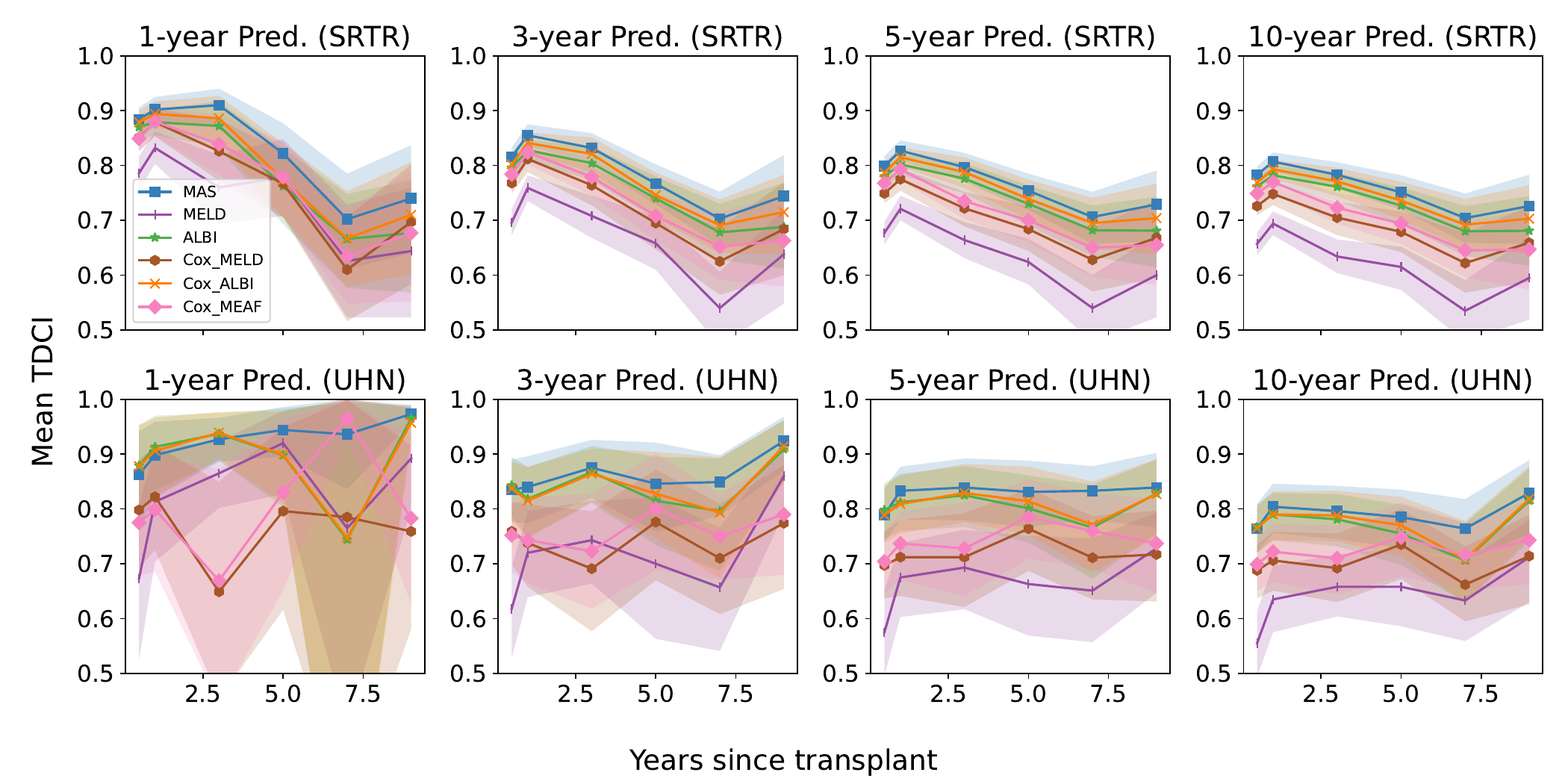}
        \caption{Model concordance on predicting 3-year, 5-year, and 10-year graft failure assessed at different prediction points. The upper row are results on the SRTR dataset, and the bottom row corresponds to the UHN dataset. We bootstrap 1000 samples to obtain the 95\% confidence intervals. This is the complete version with all risk scores plotted.}
    \label{fig:5}
\end{figure}

\end{document}